%% file: arxiv.tex
\documentclass[letterpaper, 10 pt, conference]{ieeeconf}  % Comment this line out if you need a4paper

\IEEEoverridecommandlockouts                              % This command is only needed if 
\usepackage{graphicx} % for pdf, bitmapped graphics files
\usepackage{epsfig} % for postscript graphics files
\usepackage{mathptmx} % assumes new font selection scheme installed
\usepackage{times} % assumes new font selection scheme installed
\usepackage{amsmath} % assumes amsmath package installed
\usepackage{amssymb}  % assumes amsmath package installed
\usepackage{xcolor}
\usepackage{booktabs}
\usepackage{algorithm}
\usepackage{algorithmic}
\usepackage{dblfloatfix} % more predictable placement for two-column floats
\usepackage{placeins}
\usepackage{pgfplots}
\pgfplotsset{compat=1.18}

\definecolor{dpgray}{HTML}{9E9E9E}
\definecolor{oursteal}{HTML}{009E73}

\newcommand{\runinhead}[1]{\par\noindent\textbf{#1.}\ }

\title{\LARGE \bf
Robot Learning with Visual Predicted Force
}
\author{%
Haonan Chen$^{1,*}$, Feiyang Wu$^{1,*}$, Yuxiang Ma$^{2}$, Mustafa Mete$^{1}$, Pengfei Ye$^{2}$,\\
Junxuan Shen$^{2}$, Cheng Zhu$^{3}$, Aurora Ruggeri$^{1}$, Kelvin Cheung$^{1}$,\\
Jiayuan Mao$^{3}$, Edward Adelson$^{2}$, Jiajun Wu$^{4}$, Robert D. Howe$^{1}$, Yilun Du$^{1}$%
\thanks{$^{*}$Equal contribution.}%
\thanks{$^{1}$Harvard University, Cambridge, MA, USA.}%
\thanks{$^{2}$Massachusetts Institute of Technology, Cambridge, MA, USA.}%
\thanks{$^{3}$University of Pennsylvania, Philadelphia, PA, USA.}%
\thanks{$^{4}$Stanford University, Stanford, CA, USA.}%
}

\begin{document}

\maketitle
\thispagestyle{empty}
\pagestyle{empty}

%%%%%%%%%%%%%%%%%%%%%%%%%%%%%%%%%%%%%%%%%%%%%%%%%%%%%%%%%%%%%%%%%%%%%%%%%%%%%%%%
\begin{abstract}

Force-aware manipulation typically relies on specialized force or tactile sensors. We show that force-aware manipulation can instead be achieved through visual force prediction from the deformation of a compliant Fin Ray gripper. Our approach trains two models. First, we train a visual force estimator on calibration data and use it to annotate task demonstrations with force estimates. Second, we train an action--force proposal policy on these force-augmented demonstrations to jointly generate candidate robot actions and their associated forces. At test time, we sample candidate actions and the forces they are expected to produce, then execute the action whose predicted force is closest to a target from the demonstrations. We evaluate our approach on berry picking, empty-can grasping, in-hand reorientation, and plug insertion. Our results show that visual force prediction can guide inference-time action selection for contact-rich manipulation without requiring force or tactile sensors at deployment. Project Website: https://visual-force.github.io/.

% Force feedback is essential for contact-rich robotic manipulation, but commercial force-torque and tactile sensors often require costly hardware, specialized integration, and calibration. In this work, we ask whether interaction forces can instead be inferred directly from visual observations. We propose a vision-based force estimation method that leverages visible deformation during contact to predict interaction forces without dedicated force or tactile sensing hardware.
% Our estimator is trained from paired gripper-deformation images and force measurements collected with a top-down camera and instrumented contact shells. Given a masked image of the gripper deformation, an encoder-only U-Net predicts the vertical contact force, with synthetic occlusion and visual augmentations improving robustness to deployment conditions. We further use the estimator as a test-time scaling signal for a frozen diffusion policy: multiple action chunks are sampled from the policy and re-ranked according to how well their predicted force matches a desired contact force. This enables force-aware manipulation using only visual input at deployment. Experiments demonstrate that visual force estimation can support force-aware robotic manipulation without specialized sensing hardware.

\end{abstract}

%%%%%%%%%%%%%%%%%%%%%%%%%%%%%%%%%%%%%%%%%%%%%%%%%%%%%%%%%%%%%%%%%%%%%%%%%%%%%%%%
\section{INTRODUCTION}

% Force and torque are fundamental signals in physical interaction. Humans rely on force feedback in many everyday manipulation tasks, such as assembling furniture, inserting objects, or handling fragile items. In robotics, force-torque sensing plays a similarly important role in contact-rich manipulation, where it helps detect contact, regulate interaction forces, and improve task robustness.

Controlling interaction forces is important for contact-rich manipulation,
where excessive or insufficient force can damage objects or cause task
failure. This is particularly important when grasping soft objects (e.g.,
fruits, soft packages), handling deformable objects, or performing precise
insertion~\cite{calandra2018more,qi2023inhand}. Existing robotic systems obtain such
feedback using wrist-mounted force/torque sensors~\cite{lee2019making},
flexible printed circuit board (FPCB)-based tactile
sensors~\cite{chen2026multimodal, huang20243dvitac}, or optical tactile
sensors such as GelSight Mini~\cite{gelsightmini}. However, these sensing solutions introduce hardware and integration overhead and can require changes to the manipulator's form factor and mechanical structure. Commercial six-axis
force/torque sensors from manufacturers such as ATI and Bota can cost several
thousand dollars per unit, excluding robot-specific mounting hardware and
system integration~\cite{ati_price,bota_price}. FPCB-based
tactile sensors often require custom fabrication, wiring, and calibration,
while optical tactile sensors require embedded cameras, illumination, and
compliant sensing surfaces that may wear under repeated contact. These
limitations motivate a natural question: {Can a robot estimate force directly
from visual observations and use these estimates to guide contact-rich
manipulation?}

% Existing multimodal robotic systems typically rely on wrist-mounted force/torque sensors~\cite{lee2019making}, flexible printed circuit board (FPCB)-based tactile sensors~\cite{chen2026multimodal, huang20243dvitac}, or optical tactile sensors~\cite{chen2026costream}. However, commercial six-axis force/torque sensors from manufacturers such as ATI and Bota typically cost between \$5,000--\$7,000 per unit—excluding the expenses for robot-specific mounting hardware and system integration. FPCB-based tactile sensors often require custom fabrication, wiring, and calibration, while optical tactile sensors necessitate cameras, lighting units, and flexible sensing surfaces—the latter being prone to wear or damage during repeated contact. These requirements significantly increase the cost and complexity of deploying multimodal sensing systems.

% Despite their usefulness, commercial force-torque sensors \textcolor{red}{[TODO: List the force-torque sensors]} are often expensive and require specialized hardware integration and software calibration. These practical barriers can limit their adoption, especially in low-cost or easily deployable robotic systems.

% To reduce this burden, prior work has explored customized low-cost force sensors and tactile sensing systems. While promising, these approaches still depend on dedicated hardware, fabrication effort, and nontrivial system integration. As a result, they remain difficult to scale across platforms and applications.

% Keep the teaser at the top of the right column on the first page.
\suppressfloats[t]
\begin{figure}[t]
    \centering
    \includegraphics[width=\columnwidth]{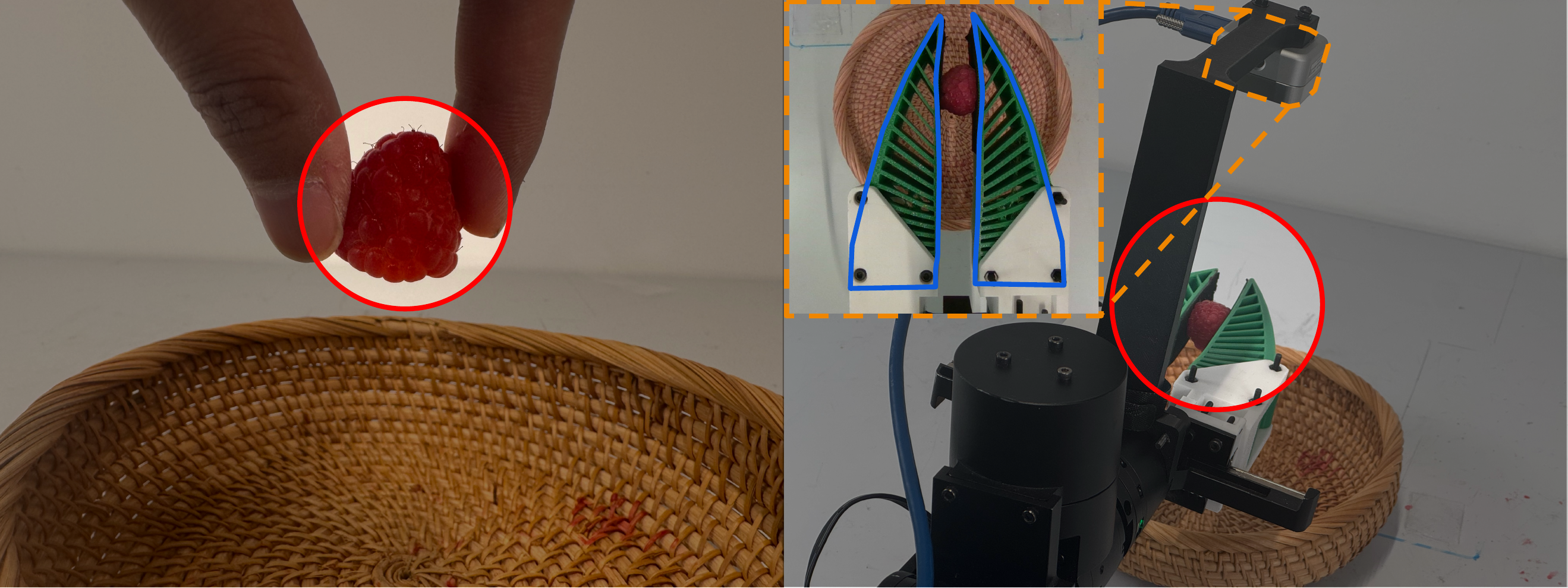}
    \caption{\textbf{Force-aware manipulation from visual gripper
    deformation.} Humans naturally regulate force when handling fragile
    objects (left). We use wrist-camera observations of a compliant gripper
    (right, inset) to infer interaction force from visible deformation and
    enable gentle robotic manipulation without dedicated force or tactile
    sensing at deployment.}
    \label{fig:teaser}
\end{figure}

In this work, we investigate learning force-aware manipulation using visual
observations of a compliant Fin Ray gripper. As illustrated in
Fig.~\ref{fig:teaser}, contact forces produce visible deformation in the
gripper, providing visual cues for force estimation. Prior studies have
explored this relationship through simulation-based force prediction and
experimental analysis of Fin Ray deformation~\cite{debarrie2021finray,kitamura2023contact}. The gripper is
inexpensive and can be fabricated by 3D printing with TPU. We train a visual
force estimator to infer force from wrist-camera images of the gripper
during contact.

% In this work, we explore a different question: can forces be estimated directly from visual observation? Our key insight is that many physical interactions leave visible traces in gripper deformation. When a rigid object contacts a deformable object, the resulting deformation often reflects the magnitude. This occurs in common manipulation settings, such as deformable grippers grasping rigid objects or rigid grippers interacting with deformable objects. If these visual deformation cues can be learned effectively, they may provide a low-cost and accessible alternative to specialized force-sensing hardware.

To use visual force estimates for manipulation, we first train a visual
force estimator on calibration data and use it to label task demonstrations.
We then train an action--force proposal policy on these demonstrations to
jointly generate candidate actions and the associated forces values from the demonstration distribution. At test time, the policy jointly samples multiple action–force candidates. We execute the action associated with the force value closest to a task-specific target derived from the demonstration force estimates. We evaluate the method on berry picking,
empty-can grasping, in-hand reorientation, and plug insertion. Our results
show that visual force estimates can guide action selection for contact-rich
manipulation without requiring force or tactile sensors at deployment.

% Motivated by this observation, we propose a method for estimating forces directly from visual observation by leveraging deformation during contact. Rather than relying on dedicated force-torque or tactile sensors, our method uses standard visual input to infer physical interaction signals, making force-aware perception more accessible for robotic manipulation (Fig.~\ref{fig:teaser}).

Our main contributions are as follows:
\begin{itemize}
    \item We introduce a visual force estimation method that infers force from the visible deformation of a compliant Fin Ray gripper.
    \item We develop an action--force proposal policy and an inference-time selection procedure that uses target forces from demonstrations to guide action selection.
    \item We demonstrate the effectiveness of the approach on four contact-rich manipulation tasks (berry picking, empty-can grasping, in-hand reorientation, and plug insertion) without requiring force or tactile sensors at deployment.
\end{itemize}

\section{RELATED WORKS}

\subsection{Hardware-Based Force and Tactile Sensing}
Dedicated force and tactile sensors obtain contact information through
several sensing mechanisms. Force-torque sensors provide accurate measurements
but require mounting, wiring, and calibration, while piezoelectric tactile
sensors capture high-frequency stimuli~\cite{zhang2022finger}. Optical tactile
sensors instead image a compliant sensing surface~\cite{shimonomura2019tactile}.
GelSight recovers contact geometry and force-related signals~\cite{yuan2017gelsight,dong2017improved},
with subsequent designs exploring compact integration and biomimetic
structures~\cite{donlon2018gelslim,lambeta2020digit,wardcherrier2018tactip}.
Related approaches reconstruct contact geometry and estimate forces from
tactile images~\cite{lin2022dtact,lin20239dtact,sferrazza2019tactile}, while
GelSight Fin Ray variants integrate optical tactile sensing into compliant
fingers~\cite{liu2022gelsight,liu2023gelsight}. Another direction observes the
gripper structure itself. De Barrie et al.~\cite{debarrie2021finray} use
finite-element simulation data to train a model that predicts contact forces
and stress distributions from finger deformation images. Kitamura et al.~\cite{kitamura2023contact}
relate camera-observed finger deformation to contact forces measured by 16
sensors embedded in the grasped object. Stuhne et al.~\cite{stuhne2022sofia}
introduce SofIA, which uses a depth camera to observe the state of its
compliant fingers. Forces for Free estimates contact force from visual
observations of a tendon-driven underactuated hand optimized for force
observability~\cite{zhu2025forcesforfree}. Building on visual observation of
compliant structures, our work uses estimated forces to label demonstrations
and train a policy whose generated force values guide action selection.

\subsection{Force and Tactile Feedback for Robotic Manipulation}
Force and tactile information support both contact perception and manipulation
control. For perception, prior work uses touch to assess and improve
grasps~\cite{calandra2017feeling,calandra2018more,lee2019making}, estimate object
pose from contact~\cite{bauza2023tac2pose}, and track object shape and pose
during manipulation~\cite{suresh2024neuralfeels}. These contact cues also
support grasp adjustment, sliding, pivoting, and rolling~\cite{tian2019manipulation,hogan2018tactile,hogan2020tactile}.
Policies combining vision and touch extend these capabilities to learned
manipulation and in-hand rotation~\cite{chen2022visuo,qi2023inhand}, with
further work incorporating audio~\cite{li2023see} or adapting vision-only
policies to force/torque sensing~\cite{clark2026multisensorycontinuallearningadapting}.
For control, tactile feedback supports cable following and needle
threading~\cite{cable_manip,yu2023precise}, while tactile-informed dynamics
support predictive control for packing~\cite{ai2024robopack}. Hierarchical
and reactive approaches integrate contact feedback into broader manipulation
systems~\cite{fazeli2019see,hogan2020reactive,chen2026costream}. Other methods
explicitly learn force or compliance control: RETAF separates tactile
grasp-force control from arm pose prediction~\cite{kang2026learning}, Adaptive
Compliance Policy learns variable compliance from demonstrations~\cite{hou2025adaptive},
and UMI-FT uses force-labeled demonstrations collected with a handheld
interface to learn motion, grasp force, and stiffness~\cite{choi2026wild}.
Our method uses visual force estimates as training labels and policy-generated
forces for inference-time action selection, without dedicated contact-sensing
hardware at deployment.

\subsection{Test-Time Searching}
Prior work has explored inference-time search over sampled robot actions using learned interaction models~\cite{chen2023predicting} and diffusion-based planning~\cite{ma2025scalingdiff,zhang2025inferencetimescalingdiffusionmodels,janner2022planning,reuss2023goal}.
Frozen generative robot policies can be improved at test time by ranking action samples with auxiliary objectives. Value-Guided Policy Steering (V-GPS)~\cite{nakamoto2024steering} ranks policy samples using an offline-learned value function. Bidirectional Decoding~\cite{liu2024bidirectional} selects action chunks using consistency with previous decisions and contrast among future plans. Generative Predictive Control (GPC)~\cite{qi2026inference} ranks and refines policy samples using a learned world model. DynaGuide~\cite{du2025dynaguidesteeringdiffusionpolices} guides generative policies with learned dynamics, VLS~\cite{liu2026vls} uses VLM-based trajectory rewards, and TouchGuide~\cite{zhang2026touchguideinferencetimesteeringvisuomotor} steers pretrained generative visuomotor policies~\cite{chi2023diffusionpolicy,lipman2023flow} by using a tactile-informed Contact Physical Model to refine early visual-policy samples toward physically feasible contact. Our method uses visual force estimates to label demonstrations for policy training. At test time, the policy jointly generates candidate actions and force values. We select the action whose policy-generated force is closest to the target. The visual force estimator does not score candidates at test time.

\section{METHOD}

\begin{figure*}[t]
    \centering
    \includegraphics[width=\textwidth]{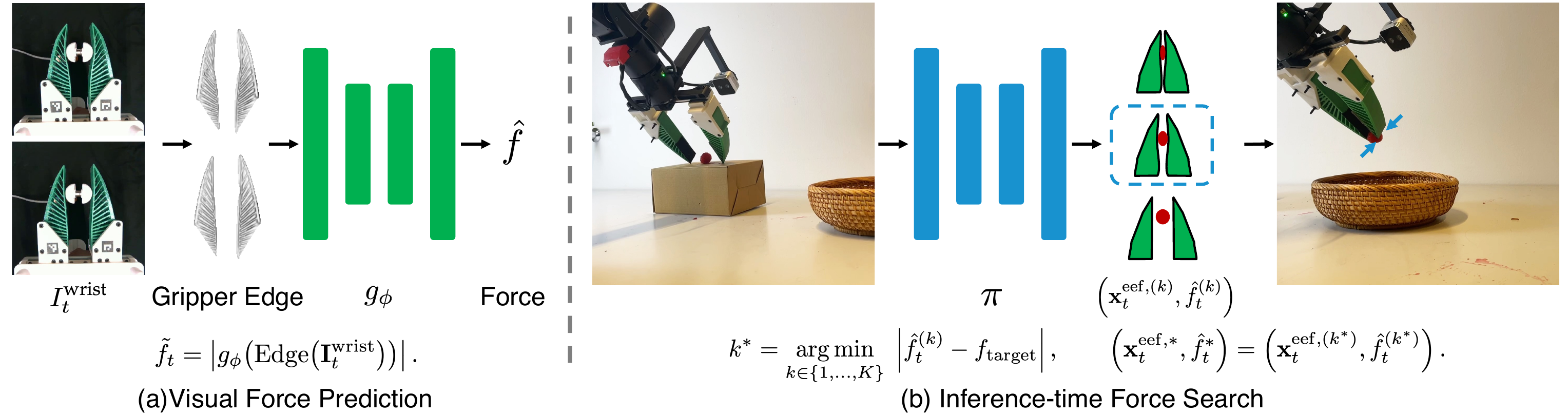}
    \caption{
    \textbf{Overview of visual force estimation and inference-time force
    search.}
    (a) Visual force estimation converts a wrist-camera image into a masked
    Sobel edge representation and uses $g_{\phi}$ to estimate the axial
    interaction force from the visible deformation of the compliant Fin Ray
    gripper.
    (b) During inference-time force search, the policy jointly samples
    candidate end-effector actions and force values. We execute the action
    from the sample whose generated force value most closely matches the
    target force.
    }
    \label{fig:method}
\end{figure*}

Our method consists of three stages. First, we collect paired images of
gripper deformation and reference force measurements to train a visual force
estimator. Second, we collect task demonstrations and use the trained
estimator to annotate them with forces and select a target force for each
task. Third, we train an action--force proposal policy on the annotated
demonstrations. At test time, we sample candidate actions and the forces they
are expected to produce, then execute the action whose predicted force is
closest to the task's target. We first formulate the problem, then describe
each stage. Fig.~\ref{fig:method} illustrates visual force estimation and
inference-time action selection.

\subsection{Problem Formulation}

We consider contact-rich manipulation using a robot equipped with a compliant
Fin Ray gripper, a wrist-mounted RGB camera, and an external side-view RGB
camera. At time $t$, the robot observes
\[
    \mathbf{o}_t
    =
    \left(
        \mathbf{I}^{\mathrm{wrist}}_t,
        \mathbf{I}^{\mathrm{side}}_t,
        \mathbf{s}_t
    \right),
\]
where $\mathbf{I}^{\mathrm{wrist}}_t$ captures the visible deformation of the
gripper, $\mathbf{I}^{\mathrm{side}}_t$ captures the global robot--object
interaction, and $\mathbf{s}_t$ denotes the robot proprioceptive state,
including the end-effector pose and gripper state.

Let $\mathbf{x}^{\mathrm{eef}}_t$ denote an end-effector action consisting of
the Cartesian position, orientation quaternion, and gripper command, and let
$f_t = |F_{z,t}|$ denote the magnitude of the axial interaction force. Given a
task-specified target force $f_{\mathrm{target}}$, our goal is to select a
task-relevant end-effector action whose associated interaction force is close
to the target.

The key challenge is that the interaction force is not directly observed
during deployment: the robot has no force/torque or tactile sensor attached to
the manipulation system. It must therefore recover force information from the
visible deformation of the compliant gripper and use this information to
select an appropriate action.

\subsection{Force Calibration and Visual Estimator Training}

We estimate the axial interaction force from the visible deformation of the
compliant Fin Ray gripper. The wrist-mounted camera provides a close-up view
of the gripper fingers, whose internal stripes and overall geometry change
under external loading. Our visual force estimator uses only this visual
deformation representation as input.

\begin{figure*}[!t]
    \centering

    \begin{minipage}[t]{4.5cm}
        \centering
        \includegraphics[height=4.5cm]{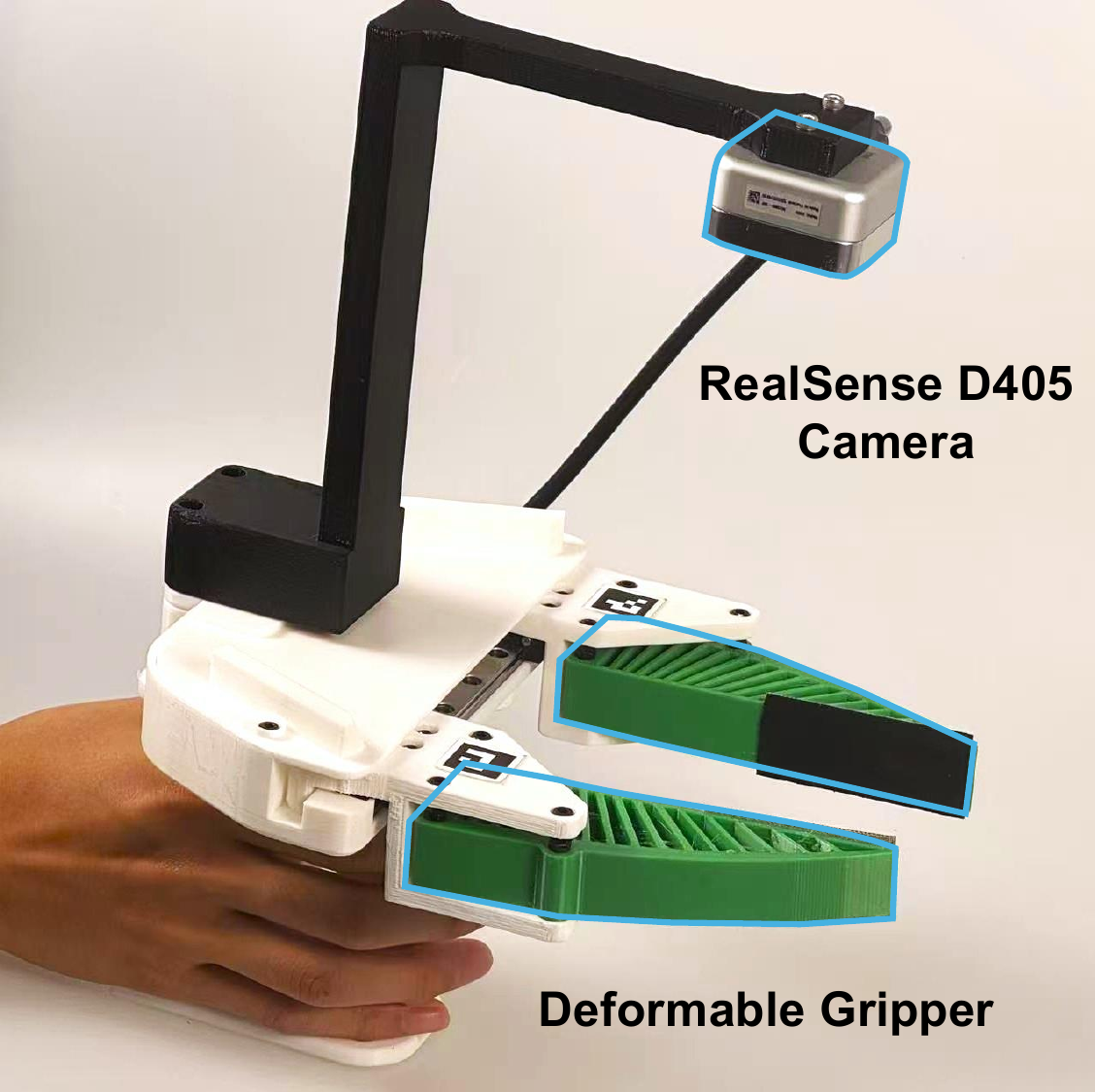}
        \par\smallskip
        {\footnotesize (a) Data collection setup.\par}
    \end{minipage}%
    \hspace{0.01cm}%
    \begin{minipage}[t]{7cm}
        \centering
        \includegraphics[height=4.5cm,pagebox=mediabox]{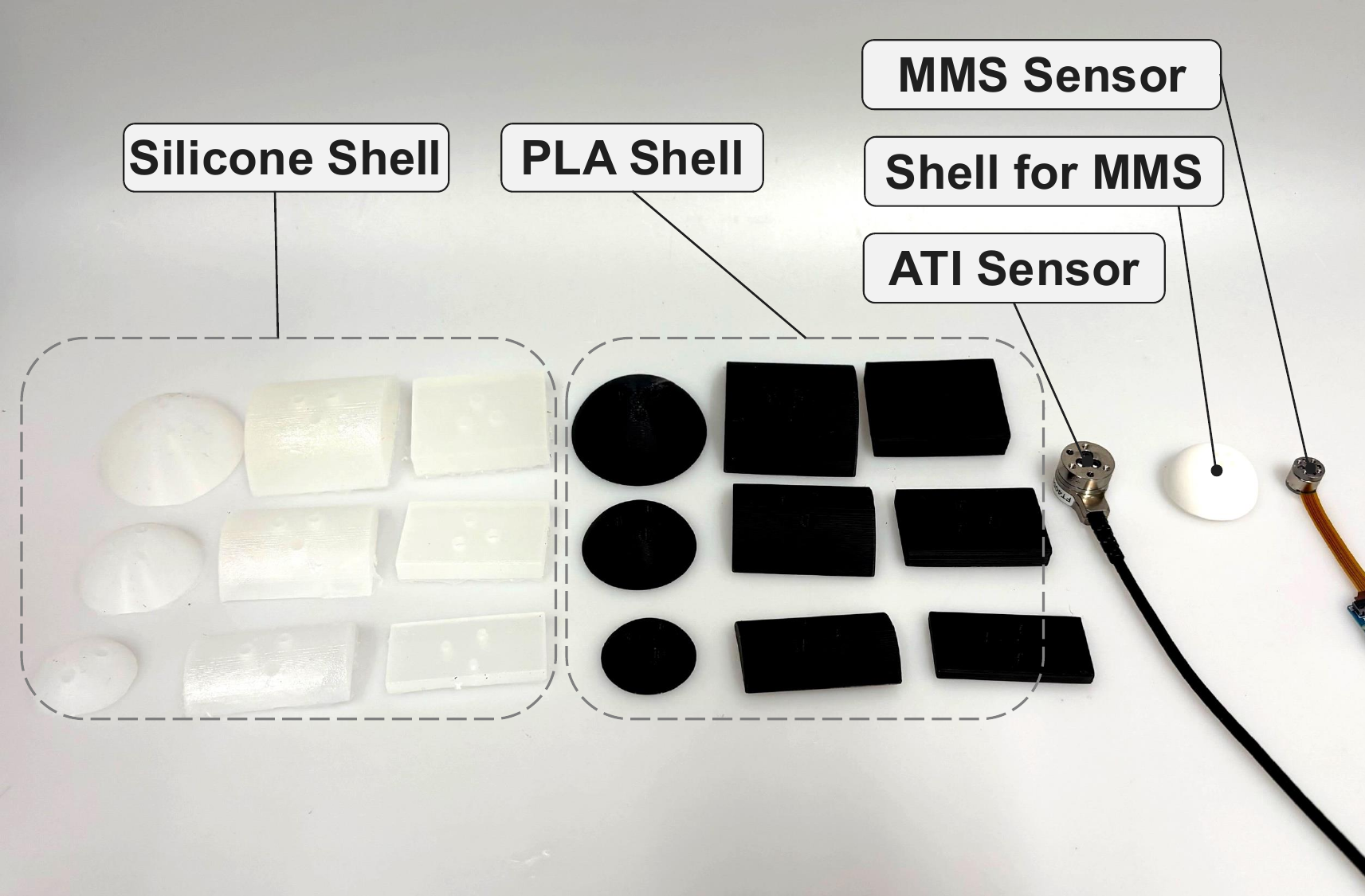}
        \par\smallskip
        {\footnotesize (b) Six-axis F/T sensor with contact fixtures.\par}
    \end{minipage}%
    \hspace{0.01cm}%
    \begin{minipage}[t]{6.0cm}
        \centering
        \includegraphics[height=4.5cm]{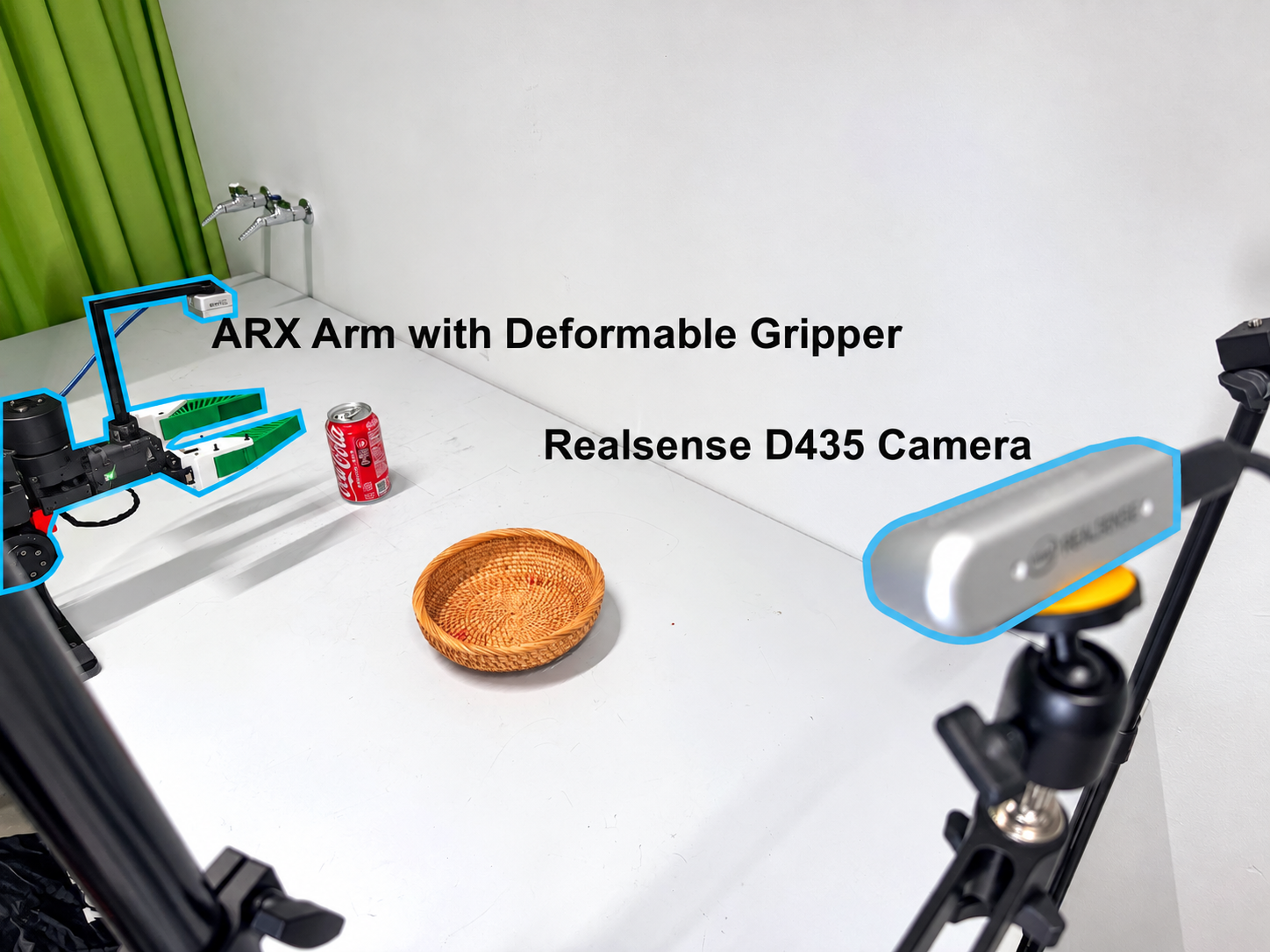}
        \par\smallskip
        {\footnotesize (c) Robotic manipulation setup.\par}
    \end{minipage}

    \caption{\textbf{Hardware setups used in our experiments.}
    (a) Visual force data collection setup, where an Intel RealSense D405
    camera observes the deformation of the hand-held Fin Ray gripper and a
    six-axis F/T sensor records the ground-truth interaction force through
    3D-printed contact fixtures.
    (b) The six-axis F/T sensor and 3D-printed contact fixtures used to record
    ground-truth interaction force during data collection.
    (c) Robotic manipulation platform used for policy evaluation, consisting
    of an ARX robotic arm equipped with wrist and side-view cameras.}
    \label{fig:hardware-setups}
\end{figure*}

\runinhead{Visual Force Data Collection}
Training the visual force estimator requires paired observations of gripper
deformation and ground-truth force measurements. To collect these pairs, we
mount 19 3D-printed contact shells
(Fig.~\ref{fig:hardware-setups}(b)) on a reference force sensor.
During contact, the fixtures transmit the force applied by the gripper to the
reference sensor, while the wrist-mounted camera records the corresponding
deformation of the Fin Ray fingers. We collect data using both a Mitsumi
MMS101 resistive force sensor and an ATI six-axis force/torque (F/T) sensor
as reference sensors.

We press the gripper against the fixtures at diverse contact locations and
with varying force magnitudes, ranging from gentle contact to firm loading.
We collect 117 episodes totaling 47.5 minutes. Images are recorded at 30 Hz
and force measurements at approximately 20 Hz. We align the streams by
timestamp and use only $F_z$ to train the force estimator.

\runinhead{Model Training and Inference}
Given a wrist-camera image $\mathbf{I}^{\mathrm{wrist}}_t$, we first use
SAM2~\cite{ravi2025sam} to segment the Fin Ray gripper. We remove the
background using the segmentation mask and resize the masked image to
$256\times256$. We then convert the masked image to grayscale and compute its
Sobel edge magnitude. To suppress artificial edges introduced by the
segmentation boundary, we erode the gripper mask by one pixel and apply the
eroded mask to the edge image. The resulting representation highlights the
internal stripes and geometry of the Fin Ray fingers while reducing
task-irrelevant variation from the object and background.

The resulting single-channel edge image is processed by an encoder-only
U-Net~\cite{DBLP:conf/miccai/RonnebergerFB15} with feature dimensions
$(32,64,128,256)$ and a 512-channel bottleneck. The network is trained by
minimizing the mean-squared error between the predicted signed axial force and
the timestamp-aligned $F_z$ measurement.

For downstream manipulation, we use the magnitude of the predicted axial
force. The complete visual force estimation pipeline is
\begin{equation}
    \tilde{f}_t
    =
    \left|
        g_{\phi}
        \left(
            \operatorname{Edge}
            \left(
                \mathbf{I}^{\mathrm{wrist}}_t
            \right)
        \right)
    \right|,
    \label{eq:visual_force}
\end{equation}
where $\operatorname{Edge}(\cdot)$ denotes the segmentation and edge-extraction
pipeline described above, and $g_{\phi}$ denotes the learned visual force
estimator. Thus, $\tilde{f}_t$ is inferred solely from the visual deformation
of the gripper and requires no F/T sensor or robot proprioceptive state at
deployment time.

\subsection{Task Demonstration Collection and Force Annotation}

For each manipulation task, we collect demonstrations using the robot arm,
recording camera observations, robot states, and demonstrated end-effector
actions. We freeze the visual force estimator $g_{\phi}$ trained in the first
stage and use Eq.~\eqref{eq:visual_force} to annotate each demonstration. At
timestep $t$, the demonstrated end-effector action
$\mathbf{x}^{\mathrm{eef}}_t$ is paired with the corresponding visual force
estimate $\tilde{f}_t$. These annotations produce force-augmented
demonstrations for policy training.

For each task, we select a target force $f_{\mathrm{target}}$ from the
estimated force values in its demonstration dataset. This task-specific
target is provided to the action-selection procedure at test time.

\subsection{Action--Force Proposal Policy and Inference-Time Selection}

We use the force-augmented demonstrations to train an action--force proposal
policy. At test time, we sample candidate action--force pairs from this policy
and select an action using the task-specific target force.

\runinhead{Joint Action--Force Policy Training}
We train a stochastic policy $\pi_{\theta}$ to jointly generate an
end-effector action and an associated force value from the current
observation:
\begin{equation}
    \left(
        \mathbf{x}^{\mathrm{eef}}_t,
        \hat{f}_t
    \right)
    \sim
    \pi_{\theta}
    \left(
        \cdot
        \mid
        \mathbf{o}_t
    \right).
    \label{eq:joint_policy}
\end{equation}
During training, the action component is supervised by the demonstrated
end-effector action, while the force component is supervised by the visual
force label $\tilde{f}_t$. The policy therefore learns the correspondence
between task-relevant actions and the interaction forces observed in the
demonstrations.

Importantly, $\pi_{\theta}$ is not an action-conditioned force estimator. The
visual force estimator $g_{\phi}$ predicts force solely from the visual
deformation of the gripper, whereas $\pi_{\theta}$ jointly generates an action
and a force value from the current observation. The generated force component
is used to rank candidate actions during inference and is not directly sent
to the robot as a control command.

\runinhead{Force Search}
Given the current observation $\mathbf{o}_t$, we independently sample $K=32$
candidate action--force pairs from Eq.~\eqref{eq:joint_policy}:
\[
    \left\{
        \left(
            \mathbf{x}^{\mathrm{eef},(k)}_t,
            \hat{f}^{(k)}_t
        \right)
    \right\}_{k=1}^{K}
    \sim
    \pi_{\theta}
    \left(
        \cdot
        \mid
        \mathbf{o}_t
    \right).
\]
Given a task-specified target force $f_{\mathrm{target}}$, we select the
candidate whose generated force component is closest to the target and execute
its corresponding end-effector action:
\begin{equation}
    k^{*}
    =
    \underset{k\in\{1,\ldots,K\}}{\arg\min}
    \left|
        \hat{f}^{(k)}_t
        -
        f_{\mathrm{target}}
    \right|,
    \qquad
    \mathbf{x}^{\mathrm{eef},*}_t
    =
    \mathbf{x}^{\mathrm{eef},(k^{*})}_t.
    \label{eq:force_search}
\end{equation}
Only the selected end-effector action
$\mathbf{x}^{\mathrm{eef},*}_t$ is sent to the robot. The generated force value
is used only for candidate selection. This procedure is repeated at each
timestep, allowing the target force to guide action selection without requiring
a force/torque or tactile sensor during deployment.

Algorithm~\ref{alg:force-search} summarizes the inference-time search
procedure.

\begin{algorithm}[t]
\caption{Inference-Time Force Search}
\label{alg:force-search}
\begin{algorithmic}[1]
\REQUIRE current observation $\mathbf{o}_t$, target force
$f_{\mathrm{target}}$, number of candidates $K$, frozen policy $\pi_\theta$
\FOR{$k = 1$ to $K$}
    \STATE sample
    $\left(
        \mathbf{x}^{\mathrm{eef},(k)}_t,
        \hat{f}^{(k)}_t
    \right)
    \sim
    \pi_\theta(\cdot \mid \mathbf{o}_t)$
\ENDFOR
\STATE
$k^{*}
\leftarrow
\displaystyle
\arg\min_{k \in \{1,\ldots,K\}}
\left|
    \hat{f}^{(k)}_t
    -
    f_{\mathrm{target}}
\right|$
\STATE
$\mathbf{x}^{\mathrm{eef},*}_t
\leftarrow
\mathbf{x}^{\mathrm{eef},(k^{*})}_t$
\STATE execute $\mathbf{x}^{\mathrm{eef},*}_t$ on the robot
\COMMENT{generated force used only for selection}
\STATE advance to timestep $t+1$ and repeat
\end{algorithmic}
\end{algorithm}

\section{Experiment}
\label{sec:experiments}

Our experiments evaluate whether visually estimated force can improve policy
performance on contact-rich manipulation tasks.

% Unlike the tangential forces that directly move an object, $F_z$ determines both the compressive load applied to the object and the maximum frictional force that can be transmitted through the finger--object contacts.

% To evaluate whether the proposed vision-based force estimator can provide useful feedback for real-world robotic manipulation, we design four tasks whose success depends critically on accurate regulation of the gripper force. Our estimator predicts a one-dimensional normal gripping force, denoted by $F_z$, which acts perpendicular to the finger contact surfaces and corresponds to the $z$-axis of our gripper coordinate frame. 

% The four tasks evaluate two complementary capabilities. Soda-can grasping and berry picking require the policy to regulate $F_z$ within a suitable range: the force must be sufficiently large to prevent the object from slipping, but sufficiently small to avoid deformation or damage. In-hand reorientation and plug insertion, in contrast, require the gripper to transmit tangential forces through friction. Because the available friction depends on the normal gripping force, successful execution requires the policy to regulate $F_z$ according to the contact state and task demands.

\subsection{Experimental Setup}
For visual force data collection, we use a hand-held Fin Ray gripper equipped with an Intel RealSense D405 camera, as shown in
Fig.~\ref{fig:hardware-setups}. Ground-truth axial-force measurements are recorded using both a Mitsumi MMS101 resistive force sensor and an ATI six-axis force/torque (F/T) sensor supporting the 3D-printed contact fixtures. The reference sensor is used only to
collect supervision for the visual force estimator; it is not
attached to the robot during manipulation-policy deployment.

For the robot manipulation experiments, we train a stochastic policy using the
Diffusion Policy architecture~\cite{chi2023diffusionpolicy}
and evaluate it on an ARX robot arm equipped with the same Fin Ray gripper
design. An Intel RealSense D405 wrist camera observes the deformation of the
gripper, while an Intel RealSense D435 side-view camera captures the global
robot--object interaction. The policy is trained and executed on a
single NVIDIA GeForce RTX 5090 GPU.
% After training the base diffusion policy, we evaluate it on a single ARX robotic arm equipped with an overhead camera and a side-view camera. The camera distance is set to match the UMI setup~\cite{Chi-RSS-24} exactly, preserving the same field of view and visual scale of the gripper as in the original UMI configuration. The complete setup is shown in Fig.~\ref{fig:hardware-setups}. The diffusion policy runs on a single NVIDIA GeForce RTX 5090 GPU.

% Queue results figures early enough for two-column placement.
\begin{figure*}[!t]
  \centering
  \definecolor{currentbaseline}{HTML}{D9700A}
  \definecolor{viewforce}{HTML}{009E73}
  \begin{tikzpicture}
    % Fixed panel coordinates align the photograph and both plotting areas.
    \node[anchor=south west, inner sep=0] at (0,0)
      {\includegraphics[width=2.3cm,height=2.3cm,trim=320bp 220bp 282bp 38bp,clip]{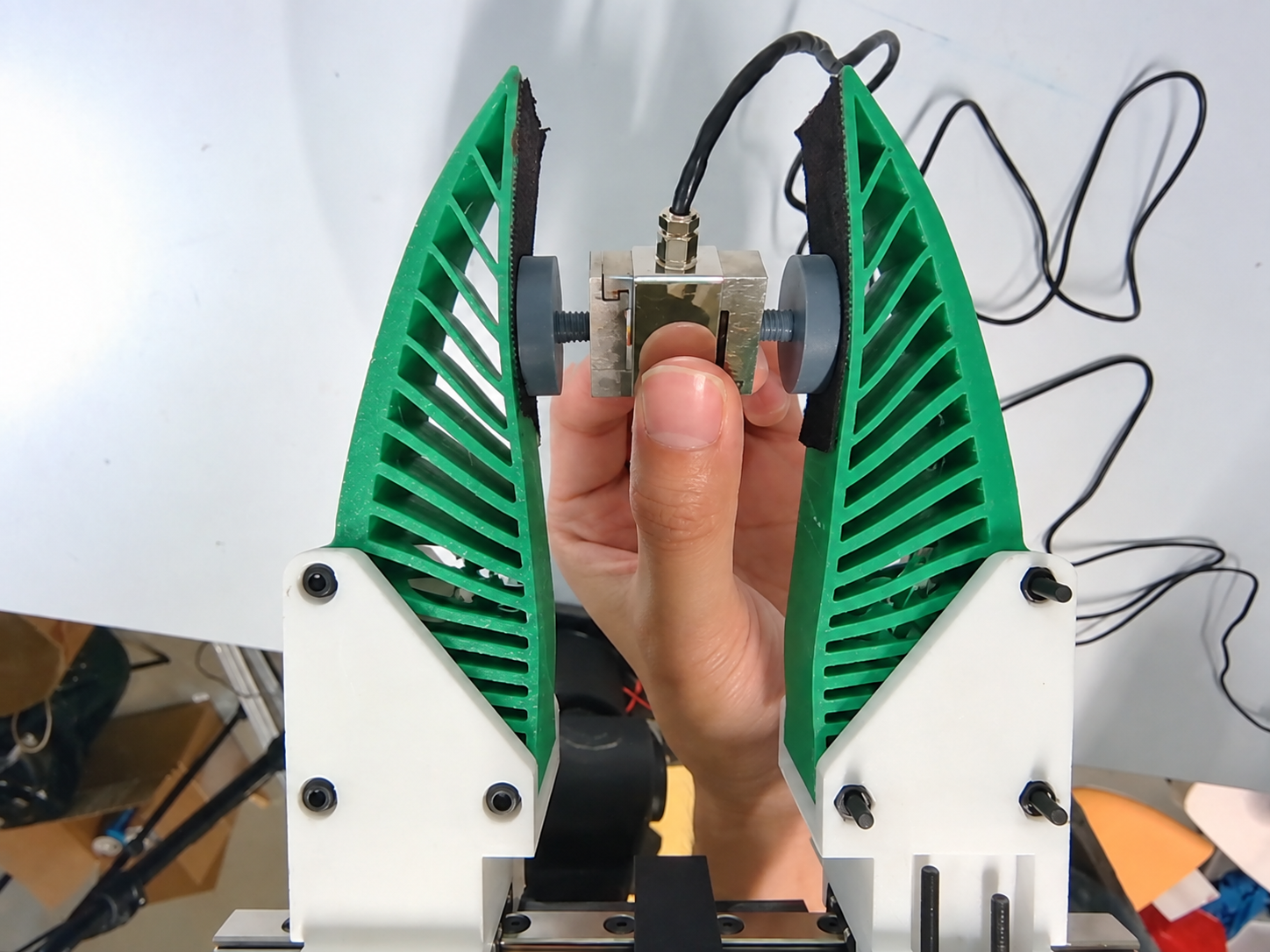}};
    \node[anchor=west, fill=white, fill opacity=0.9, text opacity=1,
          inner sep=2pt, font=\footnotesize] (sensorlabel) at (0.05,2.08) {Load cell};
    \draw[->, black!80, line width=0.65pt] (sensorlabel.east) -- (1.23,1.53);

    % One legend serves both quantitative panels.
    \matrix[anchor=south, column sep=5pt, nodes={font=\footnotesize, inner sep=0pt}]
      at (10.375,2.38) {
      \draw[black!75, dashed, line width=0.9pt] (0,0) -- (0.45,0); & \node {Reference}; &
      \draw[currentbaseline, line width=0.8pt] (0,0) -- (0.45,0); & \node {Current}; &
      \draw[viewforce, line width=1pt] (0,0) -- (0.45,0); & \node {ViewForce}; \\
    };
    \begin{axis}[
      at={(3.75cm,0)}, anchor=south west,
      width=6.5cm, height=2.3cm, scale only axis,
      xmin=0, xmax=12.8, ymin=-1, ymax=14,
      xtick={0,2,4,6,8,10,12}, ytick={0,5,10},
      xlabel={Time (s)}, ylabel={Force (N)},
      tick label style={font=\footnotesize}, label style={font=\footnotesize},
      axis lines=left, axis line style={draw=black!45, -},
      ymajorgrids=true, grid style={draw=black!10, line width=0.35pt},
      tick align=outside, tick style={draw=black!45},
    ]
      \addplot[currentbaseline, line width=0.8pt, no marks]
        table[x=time_s, y=current_linear_n, col sep=comma]{force_trajectory.csv};
      \addplot[viewforce, line width=1pt, no marks]
        table[x=time_s, y=viewforce_n, col sep=comma]{force_trajectory.csv};
      \addplot[black!75, dashed, line width=0.9pt, no marks]
        table[x=time_s, y=ground_truth_n, col sep=comma]{force_trajectory.csv};
    \end{axis}
    \begin{axis}[
      at={(11.7cm,0)}, anchor=south west,
      width=5.3cm, height=2.3cm, scale only axis,
      xmin=-2, xmax=16, ymin=-2, ymax=16,
      xtick={0,5,10,15}, ytick={0,5,10,15},
      xlabel={Measured force (N)}, ylabel={Estimated force (N)},
      tick label style={font=\footnotesize}, label style={font=\footnotesize},
      axis lines=left, axis line style={draw=black!45, -},
      xmajorgrids=true, ymajorgrids=true,
      grid style={draw=black!10, line width=0.35pt},
      tick align=outside, tick style={draw=black!45},
    ]
      \addplot[dashed, draw=black!50, line width=0.7pt, no marks]
        coordinates {(-2,-2) (16,16)};
      \node[anchor=south east, font=\footnotesize, text=black!60,
            rotate=45, inner sep=2pt] at (axis cs:12,12) {Ideal};
      \addplot[only marks, mark=square*, mark size=0.85pt,
               mark options={fill=currentbaseline, fill opacity=0.45,
                             draw=currentbaseline, draw opacity=0.45, line width=0.1pt}]
        table[x=ground_truth_n, y=current_linear_n, col sep=comma]{force_parity_points_3episodes_numeric.csv};
      \addplot[only marks, mark=*, mark size=0.85pt,
               mark options={fill=viewforce, fill opacity=0.55,
                             draw=viewforce, draw opacity=0.55, line width=0.1pt}]
        table[x=ground_truth_n, y=viewforce_n, col sep=comma]{force_parity_points_3episodes_numeric.csv};
    \end{axis}
    \node[font=\footnotesize, anchor=north] at (1.15,-0.92) {(a) Validation setup};
    \node[font=\footnotesize, anchor=north] at (7,-0.92) {(b) Force trajectory};
    \node[font=\footnotesize, anchor=north] at (14.35,-0.92) {(c) Predicted vs. measured force};
  \end{tikzpicture}
  \caption{\textbf{Force-estimation accuracy.} (a) Reference load cell.
  (b) Force trajectories. (c) Estimated versus measured force (three episodes,
  442 frames); dashed line: perfect agreement. Mean episode MAE: motor current,
  $4.20\,\mathrm{N}$; ViewForce, $0.91\,\mathrm{N}$ (equal episode weights).}
  \label{fig:force-estimation-results}
\end{figure*}
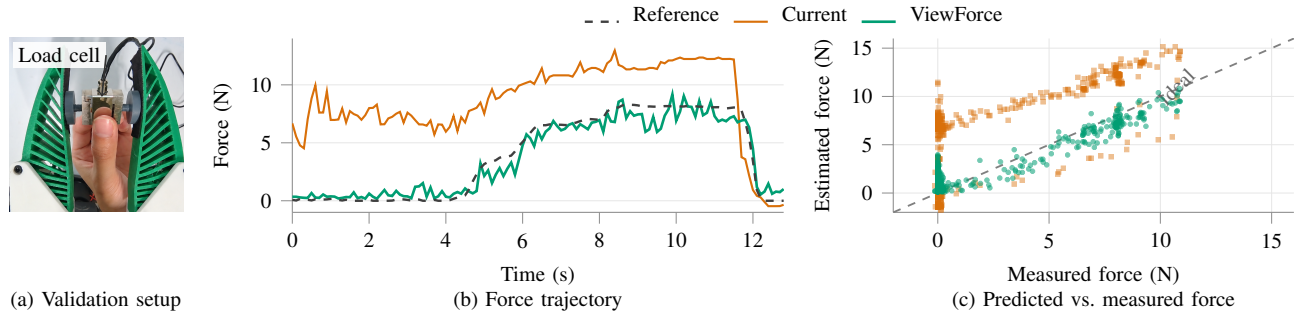

\begin{figure}[!t]
\centering
\begin{tikzpicture}
\begin{axis}[
    ybar,
    width=\columnwidth,
    height=5.4cm,
    bar width=9pt,
    enlarge x limits=0.18,
    ymin=0, ymax=112,
    ytick={0,20,40,60,80,100},
    ylabel={Success rate (\%)},
    ylabel style={font=\small},
    symbolic x coords={Soda-can, Berry, Reorient, Plug},
    xtick=data,
    xticklabel style={font=\small, align=center},
    tick align=inside,
    axis lines*=left,
    ymajorgrids=true,
    grid style={gray!25, line width=0.4pt},
    tick style={draw=none},
    legend style={
        at={(0.5,1.02)}, anchor=south,
        draw=none, fill=none, font=\small,
        legend columns=2, column sep=10pt,
    },
    legend image code/.code={
        \draw[#1] (0cm,-0.08cm) rectangle (0.24cm,0.12cm);
    },
    nodes near coords,
    nodes near coords style={font=\scriptsize, /pgf/number format/precision=1,
        /pgf/number format/fixed, /pgf/number format/fixed zerofill},
    every node near coord/.append style={yshift=1pt},
]
\addplot[draw=dpgray!75!black, fill=dpgray] coordinates {
    (Soda-can,46.7) (Berry,0) (Reorient,6.7) (Plug,26.7)
};
\addplot[draw=oursteal!80!black, fill=oursteal] coordinates {
    (Soda-can,86.7) (Berry,73.3) (Reorient,46.7) (Plug,73.3)
};
\legend{Policy (single sample), Force search (ours)}
\end{axis}
\end{tikzpicture}
\caption{\textbf{Manipulation success rates.}
Success rate (\%) over 15 trials per task for single-sample policy inference
and our force-search method, both using the same policy checkpoint.}
\label{fig:success-rates}
\end{figure}
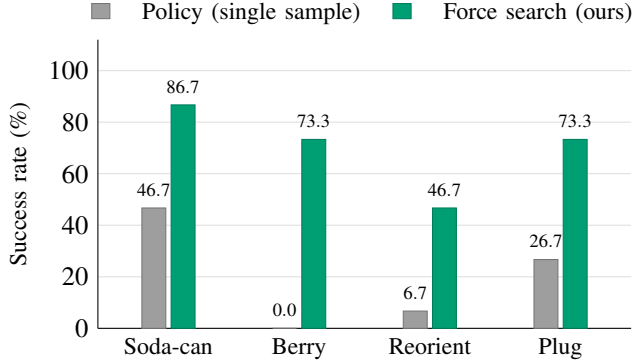

\begin{figure}[!t]
    \centering
    \includegraphics[width=0.75\columnwidth]{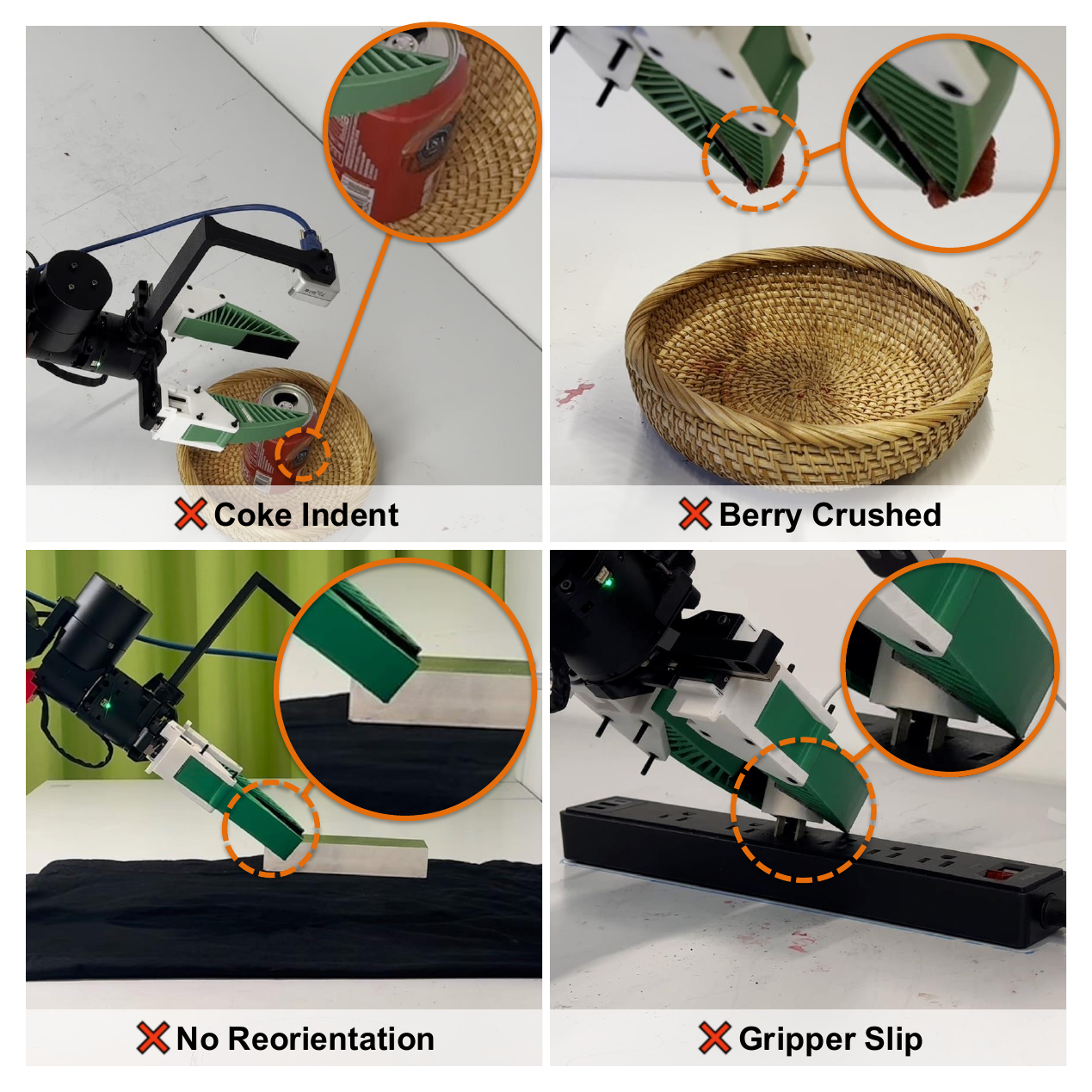}
    \caption{\textbf{Representative failure modes under single-sample policy
    inference.} From top left to bottom right: a dented empty soda can, a crushed berry,
    unsuccessful in-hand reorientation, and plug slippage within the gripper.
    These examples correspond to the four tasks reported in
    Fig.~\ref{fig:success-rates}.}
    \label{fig:baseline-failures}
\end{figure}

\begin{figure*}[!t]
    \centering
    \includegraphics[width=\textwidth]{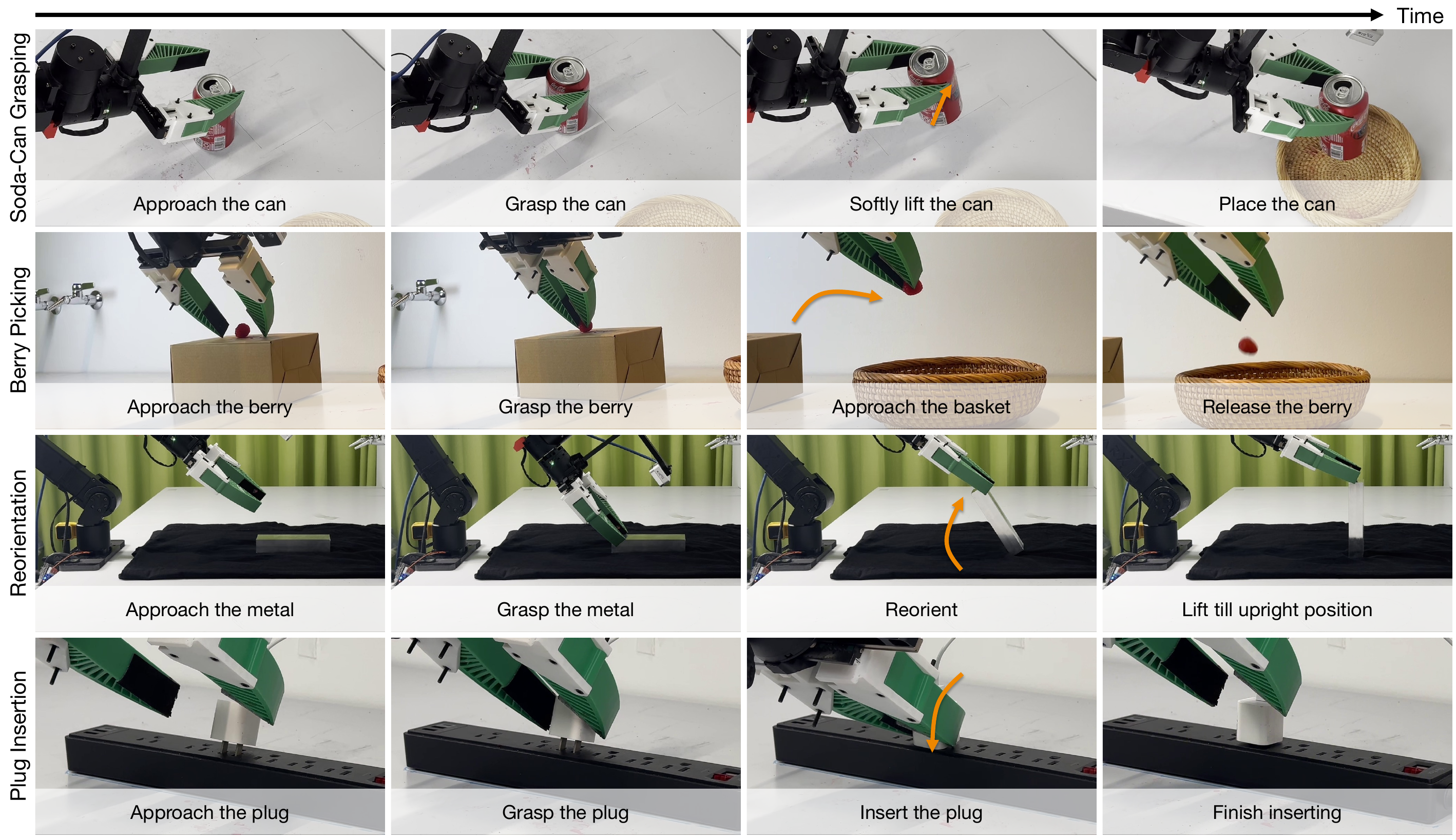}
    \caption{\textbf{Representative successful rollouts of four force-sensitive
    manipulation tasks using force-based candidate selection.} Each row
    shows a task sequence progressing from left to right. For empty soda can
    grasping and berry picking, the robot applies sufficient gripping force
    to lift and transfer the object without crushing it. During in-hand
    reorientation, the robot maintains contact while allowing the metal plate
    to pivot into an upright configuration. During plug insertion, the robot
    maintains sufficient contact friction to prevent slippage and fully
    inserts the plug into the socket.}
    \label{fig:rollout}
\end{figure*}

\subsection{Force Estimation Evaluation}
\label{sec:force_estimation_evaluation}

\runinhead{Evaluation Data}
We collect three evaluation episodes with 442 frames in total.
All three episodes are excluded from visual estimator training and
current-baseline calibration.
We record synchronized reference force measurements using a Nextech DFS-XM
force gauge with an external Micro-SBeam load cell
(Fig.~\ref{fig:force-estimation-results}(a)).

\runinhead{Compared Estimators}
We compare visual and current-based force estimation. The current baseline uses
$\hat{F} = aI + b$, where $I$ is gripper motor current. We fit $a$ and $b$
using ten separate calibration recordings. The visual estimator uses gripper
deformation as discussed previously.
%and shown in Fig.~\ref{fig:method} (a).

\runinhead{Evaluation Protocol}
We evaluate both estimators on these episodes. We compute mean absolute error
(MAE) within each episode. This is the average absolute difference between estimated
and measured force. We then average the three episode MAEs with equal weight.

\subsection{Manipulation Evaluation}
\label{sec:manipulation_experiment}

\runinhead{Experiments Design}
We evaluate our method on four force-sensitive manipulation tasks. Grasping an empty soda can
tests force regulation when handling a thin-walled object: too little force causes slippage, whereas too much force permanently deforms the can.
Berry picking additionally requires precise control at a much smaller scale,
as the berry's small size and compliance produce subtle deformation cues and
leave only a narrow margin between a stable grasp and crushing. In-hand
reorientation requires the robot to pivot a metal plate upright without
slipping or lifting it from the table, while plug insertion requires sufficient
gripping force to prevent the plug from slipping within the gripper during
insertion. Success requires transferring the can or berry without damage,
reorienting the plate upright, or fully inserting the plug.

\runinhead{Demonstration Data and Policy Training}
For each task, we collect 40 demonstrations using the ARX robot arm. We then
freeze the visual force estimator $g_\phi$ and use it to annotate every
demonstration timestep with the estimated force magnitude $\tilde{f}_t$. The
resulting action--force pairs are used to train a stochastic policy
that jointly generates an end-effector action and its associated force value.

% \begin{figure}[!t]
%     \centering
%     \includegraphics[width=\columnwidth]{figures/coke_force_analysis.pdf}
%     \caption{\textbf{Force evolution during a representative successful Soda-can
%     grasping rollout.} Columns show selected timesteps during approach, contact,
%     lift, transport, and release; rows show the corresponding side- and
%     wrist-camera views. The visually estimated normal gripping force rises
%     after contact, reaches approximately $2.0\,\mathrm{N}$ during transport,
%     and returns to nearly zero after release.}
%     \label{fig:coke-force-analysis}
% \end{figure}

\runinhead{Compared Methods}
Both procedures use the same trained policy checkpoint and differ
only in how policy samples are selected at inference time:

\begin{itemize}
    \item \textbf{Policy (single sample):}
    We draw one action--force sample from the policy, ignore its generated
    force component, and execute the corresponding end-effector action.

    \item \textbf{Ours (Force Search):}
    We draw $K=32$ action--force samples from the same policy and execute the
    action whose generated force component is closest to the task-specific
    target force.
\end{itemize}

For each task, we select a target force from the estimated force values
in its demonstration dataset.

\runinhead{Evaluation Protocol}
We evaluate each method for 15 trials on every task. Both methods are evaluated under the same initial-condition
distribution. We report the number and percentage of successful trials.

\section{RESULTS AND ANALYSIS}
\label{sec:results_analysis}

\subsection{Force-Estimation Results}
\label{sec:force_estimation_results}
We first quantify the accuracy of visually predicted forces and compare with a conventional baseline that uses gripper motor current for force estimation.
\runinhead{Estimation Accuracy}
The visual estimator achieves an MAE of $0.91\,\mathrm{N}$, compared with
$4.20\,\mathrm{N}$ for the current baseline.

\runinhead{Temporal Tracking}
Figure~\ref{fig:force-estimation-results}(b) compares the estimates over a
representative force trajectory. The visual estimate remains close to zero
during the initial low-force period. It follows the rise in measured force
and returns toward zero as the force drops near the end. The current baseline
reports substantial force even when the reference measurement is near zero.
It also overestimates force during loading. The visual estimate fluctuates at
higher forces but remains closer to the reference. Gripper deformation
therefore provides a more faithful estimate of both low-force periods and
changes during contact.

\runinhead{Prediction Accuracy}
Figure~\ref{fig:force-estimation-results}(c) compares estimated and measured
forces across all three evaluation episodes. Points on the dashed line
represent perfect predictions. Most current-based estimates lie above the
line, showing that the baseline overestimates force. It often predicts
substantial force when the measured force is near zero. Visual estimates lie
closer to the line, although some errors remain at low and high forces.
Overall, visual deformation gives more accurate force estimates than motor
current in the tested conditions.

\subsection{Force-Aware Manipulation Results}
\label{sec:manipulation_results}
We next evaluate whether force-based action selection improves success rates on four force-sensitive manipulation tasks compared with single-sample inference from the same policy.
\runinhead{Success Rates}
Figure~\ref{fig:success-rates} shows that force search improves success rates
across all four tasks: from 46.7\% to 86.7\% for can grasping, 0\% to 73.3\%
for berry picking, 6.7\% to 46.7\% for reorientation, and 26.7\% to 73.3\%
for plug insertion. Berry picking benefits most from force search, while
reorientation remains the most challenging task, reaching a success rate of 46.7\%.

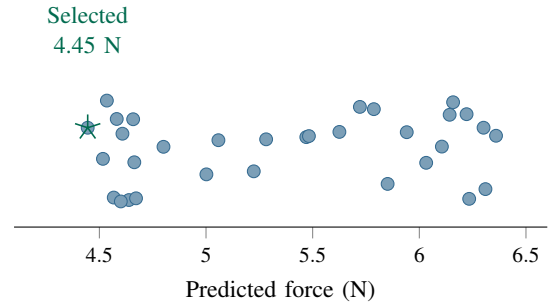
\begin{figure}[t]
    \centering
    \input{figures/soda_can_force_candidates.tex}
    \caption{\textbf{Force Search.}
    Distribution of 32 force predictions sampled from the policy for a held-out
    soda-can observation. The teal star marks the prediction selected by force
    search ($4.45\,\mathrm{N}$). Points are offset vertically for readability.}
    \label{fig:can-force-distribution}
\end{figure}

\runinhead{Force Search}
Figure~\ref{fig:can-force-distribution} illustrates how force search selects
among 32 action--force samples generated by the policy for a held-out soda-can
observation. Predicted forces range from $4.45$ to $6.36\,\mathrm{N}$, with a
mean of $5.36\,\mathrm{N}$ and a standard deviation of $0.69\,\mathrm{N}$.
Force search selects the candidate closest to the task-specific target force;
in this example, the selected candidate also has the lowest predicted force,
$4.45\,\mathrm{N}$. Single-sample inference instead executes a randomly
drawn action without considering its predicted force.

\runinhead{Qualitative Behavior}
\label{sec:qualitative_results}
Figure~\ref{fig:baseline-failures} shows failures under single-sample
inference. The can is dented and the berry is crushed. Both indicate excessive
gripping force. The plate fails to reorient, and the plug slips within the
gripper. The examples highlight the need to balance a secure grasp with the
contact conditions required by each task.

Figure~\ref{fig:rollout} shows successful rollouts with force search. The robot
lifts and transfers the can and berry without visible damage. During
reorientation, the plate pivots against the table while remaining in contact
with the gripper. During insertion, the plug stays secure as the robot pushes
it into the socket. The rollouts illustrate how suitable gripping force
supports both gentle handling and tasks that require friction.

The four tasks illustrate two roles of gripping force: balancing grasp stability
against damage, and providing friction for reorientation and insertion.
Together, the results show that visually estimated force can guide action
selection across both types of interaction.

\FloatBarrier
\section{Limitations}
% There are several limitations that motivate directions for future work.
Our work has several limitations.
First, our method requires a task-specific target force, which is obtained from successful demonstrations in our experiments. Automatically inferring and adapting the desired force based on object properties, contact state, and task phase remains an important direction for future work. Inferring object properties to parameterize adaptive grasp policies, as explored in DeliGrasp~\cite{xie2024deligrasp}, offers one possible direction for selecting task-appropriate target forces. Second, our current method estimates only the normal gripping force $F_z$ and does not recover tangential forces or torques. Extending visual force estimation to the full contact wrench would require deformable gripper with rich multi-axis deformation cues. Third, action selection relies on policy-generated force predictions without online visual force feedback, limiting correction of prediction errors and unexpected contact disturbances. Future work could incorporate this feedback into action execution and evaluate robustness under external disturbances.

\section{CONCLUSIONS}

% \addtolength{\textheight}{-12cm} % Uncomment and tune only for final-page column balancing.
                                  % on the last page of the document manually. It shortens
                                  % the textheight of the last page by a suitable amount.
                                  % This command does not take effect until the next page
                                  % so it should come on the page before the last. Make
                                  % sure that you do not shorten the textheight too much.

%%%%%%%%%%%%%%%%%%%%%%%%%%%%%%%%%%%%%%%%%%%%%%%%%%%%%%%%%%%%%%%%%%%%%%%%%%%%%%%%

In this work, we presented a force-aware manipulation approach that estimates
normal gripping force directly from the visible deformation of a soft gripper.
We use these visual force estimates to supervise a policy that
jointly generates end-effector actions and their associated force values. At
inference time, force search samples multiple action--force candidates and
selects the action whose predicted force best matches a task-specific target.
This enables force-aware action selection without dedicated force or tactile
sensors, policy updates, or changes to the robot control interface at
deployment.

Across four force-sensitive manipulation tasks, force search consistently
improves success over single-sample policy inference. These
results show that visual gripper deformation can provide a useful contact
signal not only for force estimation, but also for guiding policy inference.

More broadly, our results suggest that a robot's compliant structure can serve
as a visually observable sensing medium. We hope this perspective encourages
the development of low-cost manipulation systems that recover physical
interaction signals directly from visual observations of robot deformation.

%%%%%%%%%%%%%%%%%%%%%%%%%%%%%%%%%%%%%%%%%%%%%%%%%%%%%%%%%%%%%%%%%%%%%%%%%%%%%%%%

%%%%%%%%%%%%%%%%%%%%%%%%%%%%%%%%%%%%%%%%%%%%%%%%%%%%%%%%%%%%%%%%%%%%%%%%%%%%%%%%
% \section*{APPENDIX}

% \section*{ACKNOWLEDGMENT}

%%%%%%%%%%%%%%%%%%%%%%%%%%%%%%%%%%%%%%%%%%%%%%%%%%%%%%%%%%%%%%%%%%%%%%%%%%%%%%%%

\bibliographystyle{IEEEtran}
\bibliography{references}

\end{document}

%% file: figures/soda_can_force_candidates.tex
% Requires pgfplots. Compile from the repository root.
\begingroup
% Font controls: axis label, ticks, and annotations.
\newcommand{\forceLabelFont}{\fontsize{9}{11}\selectfont}
\newcommand{\forceTickFont}{\fontsize{8}{10}\selectfont}
\newcommand{\forceAnnotationFont}{\fontsize{9}{11}\selectfont}
\definecolor{candidateBlue}{HTML}{356B91}
\definecolor{selectedTeal}{HTML}{009E73}
\definecolor{selectedOutline}{HTML}{006B4F}
\pgfplotstableread[col sep=comma]{figures/force_distribution.csv}\forceCandidates
\begin{tikzpicture}
\begin{axis}[
    width=\linewidth, height=4.6cm,
    xlabel={Predicted force (N)},
    label style={font=\forceLabelFont},
    tick label style={font=\forceTickFont},
    xmin=4.1, xmax=6.6, ymin=-0.18, ymax=0.36,
    xtick={4.5,5,5.5,6,6.5}, ytick=\empty,
    axis x line*=bottom, axis y line=none,
    tick align=outside, scaled ticks=false,
    clip=false,
]
\addplot[only marks,mark=*,mark size=2.4pt,
    mark options={fill=candidateBlue!65,draw=candidateBlue}]
    table[x=force_proxy_n,y=jitter] {\forceCandidates};
\addplot[only marks,mark=star,mark size=5pt,
    mark options={fill=selectedTeal,draw=selectedOutline,line width=0.6pt},
    restrict expr to domain={\thisrow{selected}}{1:1}]
    table[x=force_proxy_n,y=jitter] {\forceCandidates};
\node[anchor=south,font=\forceAnnotationFont,text=selectedOutline,align=center]
    at (axis cs:4.445268630981445,0.21) {Selected\\4.45 N};
\end{axis}
\end{tikzpicture}
\endgroup